\documentclass[runningheads]{llncs}

\usepackage{eccv}

\usepackage{eccvabbrv}
\usepackage{anyfontsize}

\usepackage{graphicx}
\usepackage{booktabs}

\usepackage[accsupp]{axessibility}  %

\usepackage[width=122mm,left=12mm,paperwidth=146mm,height=193mm,top=12mm,paperheight=217mm]{geometry}

\usepackage{hyperref}

\usepackage{orcidlink}

\newcommand{\datasetname}[1]{Space3D-Bench}
\newcommand{\baselinename}[1]{RAG3D-Chat}

\newenvironment{tight_itemize}{
\begin{itemize}
  \setlength{\itemsep}{0pt}
  \setlength{\parskip}{0pt}
}{\end{itemize}}
\newcommand{\PAR}[1]{\vskip4pt \noindent{\bf #1~}}

\begin{document}

\title{Space3D-Bench: Spatial 3D Question Answering Benchmark} 

\titlerunning{Space3D-Bench}

\author{Emilia Szyma{\'n}ska\inst{1*} \and
Mihai Dusmanu\inst{2} \and
Jan-Willem Buurlage\inst{2} \and
Mahdi Rad\inst{2} \and
Marc Pollefeys\inst{1,2}
}

\authorrunning{E.~Szyma{\'n}ska et al.}

\institute{ETH Z{\"u}rich, Z{\"u}rich, Switzerland \and
Microsoft, Z{\"u}rich, Switzerland \\
{\fontsize{8}{14}\selectfont *Work done at Microsoft for a Master thesis.}
}

\maketitle
\vspace{-13pt}
\noindent\begin{center}
{\fontsize{8}{14}\selectfont \href{https://space3d-bench.github.io/}{https://space3d-bench.github.io/}}
\end{center}

\begin{figure}[h]
   \centering
   \vspace{-20pt}
   \includegraphics[width=\linewidth]{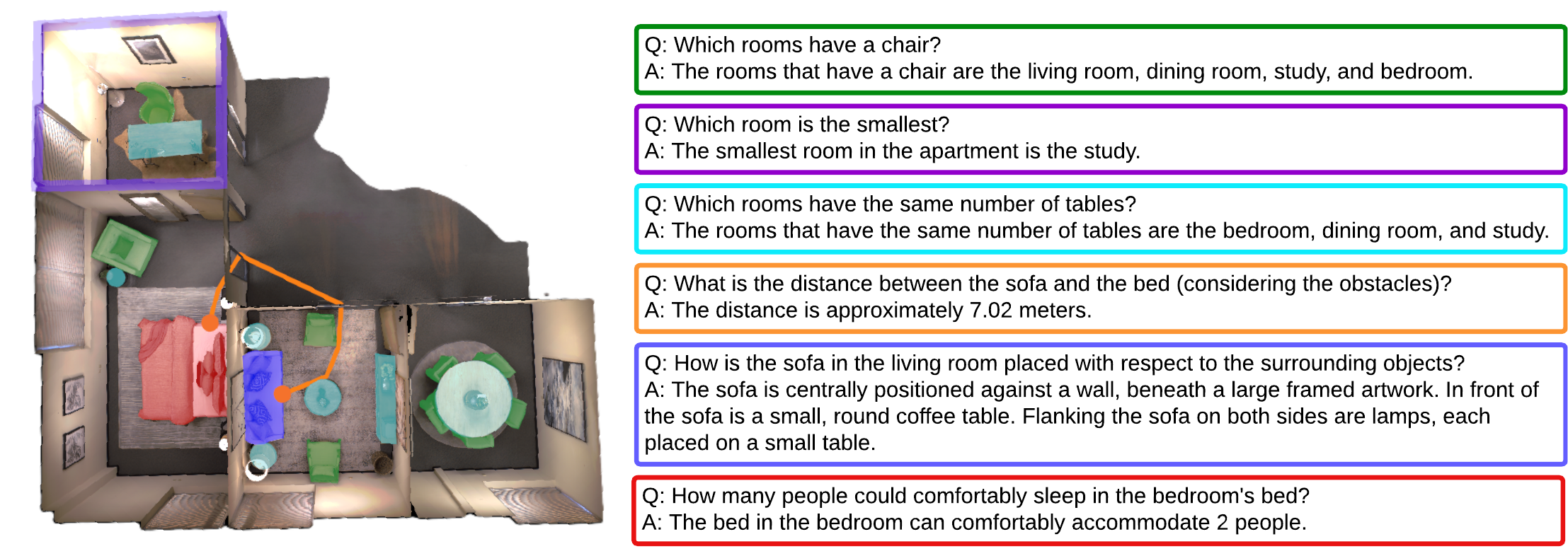}
   \vspace{-20pt}
   \caption{\textbf{Questions from \datasetname{DATASET NAME} with answers by \baselinename{the baseline}}. The dataset supports a variety of spatial tasks, including \textcolor{teal}{object location}, \textcolor{violet}{measurements}, \textcolor{cyan}{pattern identification}, \textcolor{orange}{navigation}, \textcolor{blue}{spatial relationships}, and \textcolor{red}{predictions}.}
   \vspace{-35pt}
   \label{fig:teaser}
 \end{figure}
\vspace{5pt}
\begin{abstract}
    Answering questions about the spatial properties of the environment poses challenges for existing language and vision foundation models due to a lack of understanding of the 3D world notably in terms of relationships between objects. To push the field forward, multiple 3D Q\&A datasets were proposed which, overall, provide a variety of questions, but  they individually focus on particular aspects of 3D reasoning or are limited in terms of data modalities. To address this, we present \datasetname{the dataset} -- a collection of $1000$ general spatial questions and answers related to scenes of the Replica dataset which offers a variety of data modalities: point clouds, posed RGB-D images, navigation meshes and 3D object detections. To ensure that the questions cover a wide range of 3D objectives, we propose an indoor spatial questions taxonomy inspired by geographic information systems and use it to balance the dataset. Moreover, we provide an assessment system that grades natural language responses based on predefined ground-truth answers by leveraging a Vision Language Model's comprehension of both text and images to compare the responses with ground-truth textual information or relevant visual data. Finally, we introduce a baseline called \baselinename{} integrating the world understanding of foundation models with rich context retrieval, achieving an accuracy of 67\% on the proposed dataset.
    \vspace{-12pt}

  \keywords{Spatial Q\&A Benchmark \and Retrieval-Augmented Generation \and Vision Language Model}
\end{abstract}

\vspace{-10pt}
\section{Introduction}\label{sec:intro}
Foundation models are becoming a standard tool in the field of artificial intelligence by providing adaptability and a high level of performance in a variety of down-stream tasks.
Notably, the field of 3D scene understanding has experienced rapid advancements fueled by these large pre-trained models.
More and more applications are emerging in terms of 3D reasoning, spatial awareness, and environment interaction~\cite{3dllm2024ma}. These spatial competences have provided significant improvements for mixed reality~\cite{fang2024enablingwaypointgenerationcollaborative, delatorre2024llmrrealtimepromptinginteractive, Bozkir_2024}, robotics~\cite{wang2024largelanguagemodelsrobotics, brohan2023rt2visionlanguageactionmodelstransfer, chang2023contextaware}, autonomous vehicles~\cite{mao2023gptdriverlearningdrivegpt, mao2024languageagentautonomousdriving, cho2024languageimagemodels3dunderstanding}, inclusive technologies~\cite{chatgpt4good}, or navigation~\cite{zhou2023navgpt, navigation2018anderson, Gu_2022}.
    
    Numerous approaches have been proposed by the research community to address these 3D tasks. As these foundation models proved to be capable of understanding other modalities than text, one group of solutions involves proposing multi-modal models, integrating images~\cite{gu2023conceptgraphs}, videos~\cite{zhang2020doesexistspatiotemporalvideo}, or 3D data such as point clouds or meshes~\cite{xu2023pointllm}. Another strong trend in the field is combining the strengths of existing models with such tools as context retrieval~\cite{ning2024llmfindautonomousgisagent} or zero-shot learning~\cite{zhang2024agent3dzeroagentzeroshot3d, Yuan_2024_CVPR}. The progress in understanding the spatial properties is evident, however, robustness and alignment still remain a challenge.
    
    To measure the performance of these spatially-aware systems, multiple 3D Question and Answer (Q\&A) datasets have been proposed~\cite{azuma_2022_CVPR, Zhu_2023_ICCV, ma2022sqa3d, chen2020scanrefer, yan2023comprehensive, li2023m3dbench}. They vary in their assessment objective, size, scene types, and provided data. Although, from the holistic perspective, the datasets provide a large variety, individually they are either limited in terms of accompanying data modalities, focus on a narrow aspect of 3D reasoning, or do not have a balanced question distribution in terms of objectives.
    
    To address these limitations, we built a dataset composed of $1000$ questions with ground truth answers.
    To assure that the questions cover a wide range of 3D objectives, we adapted an existing taxonomy of spatial question~\cite{understanding2024schmidts} used in Geographic Information Systems and adapted it to the indoor scenes scenario.
    We balanced the number of questions with respect to the presented categories. The questions are associated with thirteen selected scenes from the Replica dataset~\cite{replica19arxiv}, that gives access to a variety of data, such as 3D object detections, navigation meshes, and point clouds. Additionally, as Replica is integrated into Habitat Sim's environment~\cite{habitat19iccv}, data such as videos, and RGB-D or semantically-segmented images with camera poses can be seamlessly extracted.

    To complement the dataset's functionality, we developed a Vision Language Model (VLM) based automatic assessment system that evaluates the responses from a question answering system against the dataset's ground truth. To establish the assessment's correctness and reliability, we conduct an extensive user study of $60$ participants, on a subset of $40$ questions that are randomly sampled.
    As the result of the survey shows, our evaluation system agrees on 97.5\% of the cases with users, which confirms the reliability of our evaluation protocol. 

    To demonstrate a baseline performance on the created dataset, we propose \baselinename{NAME} -- Retrieval Augmented Generation (RAG) for 3D Chat -- a system that utilizes RAG~\cite{rag2020} and VLMs to identify the relevant scene context from images, texts, and an SQL database, and also has the capability to answer questions regarding navigable distances. We employed a planner based on a Large Language Model (LLM) to chain available functionalities of the system to answer complex questions. This system scored 67\% of accuracy on the dataset, which proves that there is room for improvement in the robustness of 3D spatial Q\&A. Selected questions from Space3D-Bench dataset, answered by RAG3D-Chat, are presented in \cref{fig:teaser}.

    In summary, this paper introduces the following contributions:
    \begin{tight_itemize}
        \item We propose a dataset of $1000$ diverse spatial questions and answers, based on the scenes of the Replica dataset, which offer a variety of data modalities. We present an application of a geographic spatial questions taxonomy to indoor scenes, and balance our questions accordingly.
        \item We provide a VLM-based assessment system that evaluates natural language responses given ground truth answers. To confirm the reliability of our proposed system, we conducted an extensive user study.
        \item We leverage the strengths of foundation models, and combine them with Retrieval-Augmented Generation, to present a baseline achieving $67\%$ of accuracy on the proposed dataset.
        \item We release the dataset with the assessment system, to encourage the research community to address the challenges of 3D question answering by developing and evaluating their spatial Q\&A systems.
    \end{tight_itemize}

\section{Literature Overview}

    \textbf{Spatial Q\&A Benchmarks and Datasets.}
        A variety of spatial Q\&A datasets with associated benchmarks have been constructed to tackle different aspects of spatial question answering. These datasets collectively offer diverse modalities. SpartQA~\cite{mirzaee-etal-2021-spartqa}, for example, provides textual stories describing scenes with 2D geometrical figures, based on which questions regarding spatial relationships are asked. However, since the research community extended the applications of foundation models beyond text, more complex, real-life scenes were included into Q\&As. Multi-view images paired with questions in 3DMV-VQA~\cite{hong2023threedclr} allow for evaluation of a system's abilities with respect to object counting and existence, relations and comparisons. M3DBench~\cite{li2023m3dbench} interleaves modalities in instruction-response pairs, combining texts, coordinates, images, and 3D objects, and thus offering a promising benchmark for general multi-task systems.

        ScanNet~\cite{dai2017scannet}, giving access to posed RGB-D image sequences, surface reconstructions and instance-level semantic segmentations, has been selected as a source of indoor scene context for some Q\&A datasets. SQA-3D~\cite{ma2022sqa3d} based its question on egocentric situation awareness within ScanNet's environments. Azuma's \etal ScanQA~\cite{azuma_2022_CVPR} introduced a task -- with a corresponding dataset -- of combining the answer on 3D scans with 3D bounding boxes. Simultaneously-developed Ye's \etal ScanQA~\cite{ye20243dQA} leveraged human annotations to correlate the scenes with questions and free-form answers. ScanScribe~\cite{Zhu_2023_ICCV}, although inherently provides only descriptions for indoor scenes from both ScanNet and 3RScan~\cite{Wald2019RIO}, proved to be useful for pre-training question answering models.
        
        The aforementioned 3RScan similarly  offers posed RGB-D sequences, instance-level semantic segmentation, object alignment and 3D meshes for indoor spaces, other datasets were also based upon it. One example of such is CLEVR3D~\cite{yan2023comprehensive}, covering Q\&A in the aspects of objects' attributes and their spatial relationships.
        
        Some benchmarks aim to provide an extensive evaluation on approaching specific tasks by combining various datasets. LAMM~\cite{yin2023lammlanguageassistedmultimodalinstructiontuning} integrates 3 point-cloud-related datasets to provide a comprehensive benchmark for 3D tasks, including 3D question answering. LV3D~\cite{cho2024languageimagemodels3dunderstanding}, on the other hand, focuses on multi-turn Q\&A, fusing fifteen 2D and 3D object recognition datasets.

        Not only indoor spaces are addressed in spatial Q\&As. NuScenes-QA~\cite{Qian2023NuScenesQAAM}, for instance, addresses visual question answering in autonomous driving by proposing question-answer pairs based upon outdoor environments from nuScenes~\cite{Caesar2019nuScenesAM}.

        Although the existing datasets are significant, we found it missing that the Q\&A datasets complemented with several modalities did not contain certain categories of questions (such as making predictions, calculating navigable distances), or did not balance questions distribution with respect to their objectives. Additionally, the majority of large datasets was automatically generated, which at times resulted in questions lacking coherence. We address this with human-formulated Q\&A, balanced among categories suggested in spatial geography research.
        
    \noindent \textbf{Spatial Questions Answering.} A multitude of approaches have been experimented with in order to solve spatial Q\&A challenge. 3D Concept Learning and Reasoning (3D-CLR)~\cite{hong2023threedclr} combines neural fields with 2D-pre-trained vision-language models and neural reasoning operators to answer questions based on multi-view images of a scene. Azuma \etal proposed a fused descriptor, ScanQA~\cite{azuma_2022_CVPR}, that links language expressions to the 3D scan's geometric features, enabling the regression of 3D bounding boxes to determine the objects described in the questions. In 3DQA-TR~\cite{ye20243dQA}, a language tokenizer is employed for question embedding, with two encoders extracting appearance and geometry information, to finally fuse modalities with 3D-L BERT to answer a question. Another integration of BERT is present in GPT4Point~\cite{GPT4Point}, where a Point-Q-Former~\cite{devlin-etal-2019-bert} aligns point-text feature, later to be analysed by a Language Model enhancing the model's ability to infer text. 3D-VisTA~\cite{Zhu_2023_ICCV}, on the other hand, leverages self-supervised pre-training via masked language/object modeling to effectively learn the alignment between texts and point clouds. LAMM~\cite{yin2023lammlanguageassistedmultimodalinstructiontuning}, being a multi-modality language model, encodes each modality by a corresponding pre-trained encoder, followed by a trainable projection layer and LoRA parameters~\cite{hu2022lora}, to eventually be decoded by a shared LLM. PointLLM~\cite{xu2023pointllm} as well uses a pre-trained encoder, although only on point clouds, whose extracted features are used by a pre-trained LLM for reasoning and generating responses. An intriguing integration of scene graphs is present in TransVQA3D~\cite{yan2023comprehensive}, which applies a cross-modal Transformer to fuse the features of language and object, to later include the scene graph initialization and conduct scene graph aware attention. NuScenes-QA baseline framework~\cite{Qian2023NuScenesQAAM} processes multi-view images and point clouds to obtain Bird's-Eye-View features, crops objects embeddings (based on the detected 3D bounding boxes), and forwards these features to a transformer-based Q\&A model. Cube-LLM~\cite{cho2024languageimagemodels3dunderstanding} applies changes to a multi-modal LLM by replacing a visual encoder, finetuning it on specific datasets, and using different resolutions and normalization techniques for inputs to enhance performance in 3D-related tasks. Finally, M3DBench's baseline~\cite{li2023m3dbench} utilizes a scene perceiver to extract scene tokens from 3D visual input, encodes multi-modal instructions into instruction tokens, which are then concatenated and fed into a frozen LLM, which generates the corresponding responses subsequently.

    The performances of the systems vary, depending on the end-task and available modalities. We propose another family of solutions for spatial problems based on Retrieval-Augmented Generation and evaluate their effectiveness on the created dataset. To the best of our knowledge, spatial question answering for indoor spaces has not been addressed with RAG-based approaches yet.

\vspace{-10pt}
\section{Benchmark}
    In this paper, we introduce \datasetname{}, a dataset composed of $1000$ spatial questions with ground truth answers for the Replica dataset scenes~\cite{replica19arxiv}. We improve Replica's object detections by correcting class name labelling, removing irrelevant objects, adding object-to-room assignments, and moving coordinates from Habitat's to Replica's coordinate system. We also provide data on rooms' centers and sizes, as well as framework-independent navigation meshes cleared of artifacts. To automatically assess the correctness of the answers, we propose an evaluation system based on both text and vision language models.
    
    \vspace{-8pt}
    \subsection{Dataset}

        Our Q\&A dataset is based on thirteen scenes of the Replica dataset~\cite{replica19arxiv}: six multi-room scenes -- a 2-floor house (\textit{apartment 0}), three multi-room apartments (\textit{apartment 1, 2}, \textit{hotel 0}), two different setups of the FRL apartment (\textit{FRL apartment 0, 1}); and seven single-room scenes -- three apartment rooms (\textit{room 0, 1, 2}), and four office rooms (\textit{office 0, 2, 3, 4}). As all FRL apartments vary only in the object distribution, it seemed unnecessary to use more than two similar scenes. We did not create questions for \textit{office 1} due to the small size of the room, the lack of visible details, and the overexposure of available frames.

\begin{figure}[tb]
           \centering
           \includegraphics[width=\textwidth]{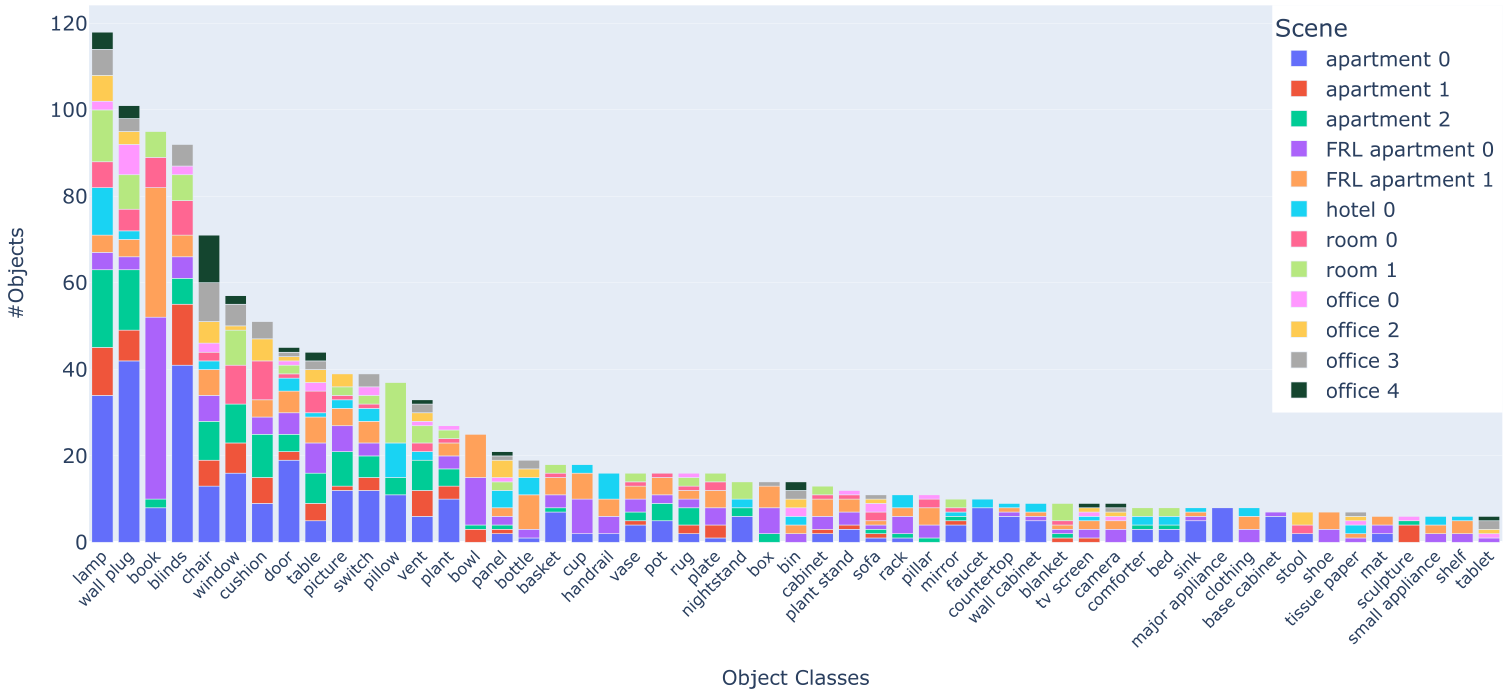}
           \vspace{-20pt}
           \caption{{\bf Distribution of detected objects across selected scenes.} Only the object classes that appear more than 6 times are included in this figure.}
           \vspace{-20pt}
           \label{fig:object_distribution}
         \end{figure}

        The distribution of detected objects with respect to their classes and corresponding scenes is presented in \cref{fig:object_distribution}. Certain objects, like ceiling lamps, wall plugs, books, or blinds, frequently appear in scenes and can be challenging for the system to distinguish due to their intra-class visual similarities. As a result, questions that would require identifying specific instances of these objects were not included in the dataset.

        We prepare 100 questions with answers for all multi-room scenes,  60 questions for apartment rooms and two offices, and 50 questions for the remaining two offices, summing up to a total of $1000$ questions with ground-truth answers. These answers may have one of the two forms: ground truth information for factual data, such as the number of objects in a room, or an illustrative image of the objects / rooms of interest for questions that involve descriptions, predictions or identifying similarities. This distinction allows to not penalize the answering system's creativity.

        Following the patterns described in \cref{sec:taxonomy}, we manually constructed both questions and answers. We decided to not generate them automatically to limit ambiguities and to ensure the suitability of questions to the scene.

\begin{figure}[tb]
        \centering
        \begin{subfigure}{0.45\textwidth}
            \includegraphics[width=\textwidth]{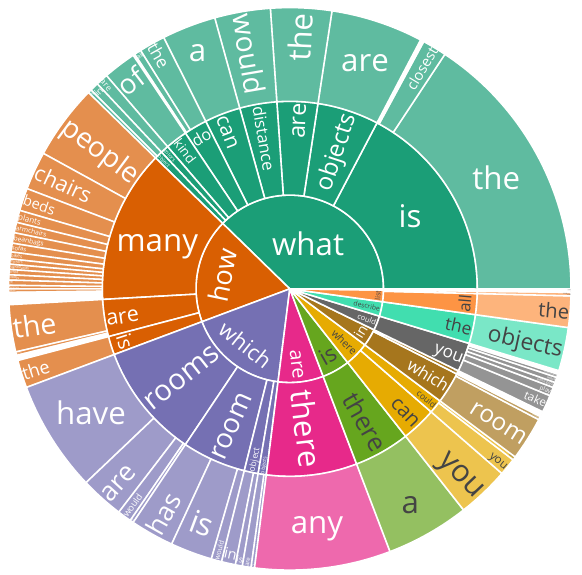}
            \caption{Question distribution based on their first three words.}
            \label{fig:sunburst}
        \end{subfigure}
        ~
        \begin{subfigure}{0.43\textwidth}
            \includegraphics[width=\textwidth]{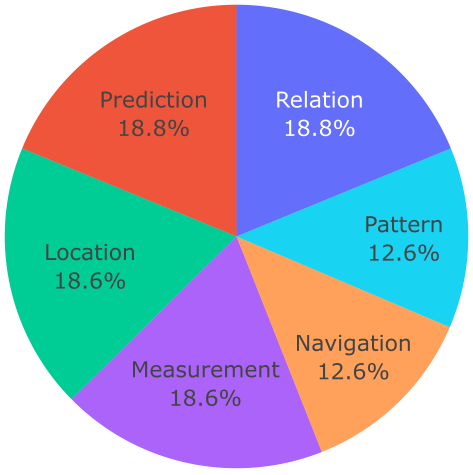}
            \caption{Question distribution based on their categories.}
            \label{fig:categories}
        \end{subfigure}
        \\
        \begin{subfigure}{\textwidth}
            \includegraphics[width=0.95\textwidth]{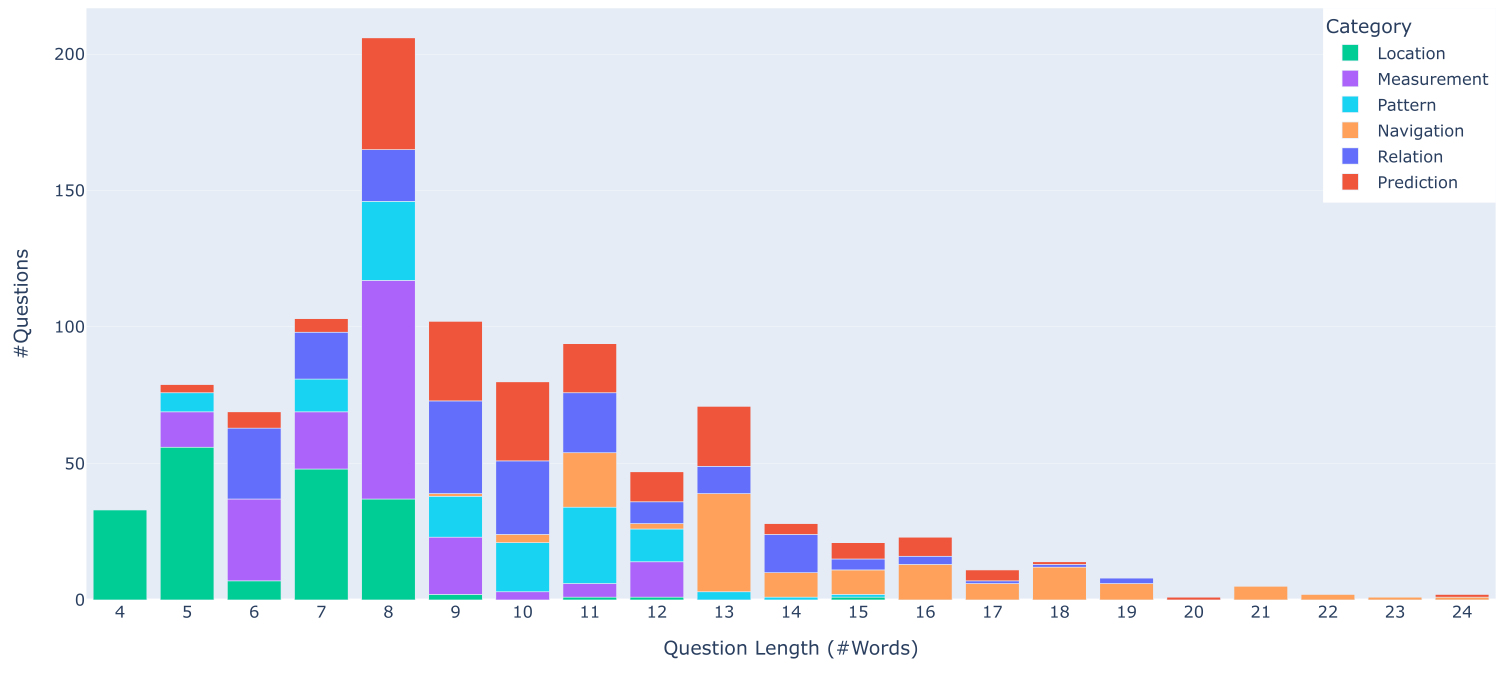}
            \caption{Question distribution based on their lengths.}
            \label{fig:lengths}
        \end{subfigure}
        \vspace{-15pt}
        \caption{{\bf Statistics of questions in our dataset.} The dataset has a large variety of phrasings (a) and is well distributed across the question categories (b). Furthermore, the questions are overall concise with an average length of around 8 words, but some longer ones are also present (c).}
        \label{fig:dataset_statistics}
        \vspace{-20pt}
        \end{figure}

        The distribution of questions with respect to the first words used to formulate them, their assigned categories, and their lengths, are presented in \cref{fig:dataset_statistics}. In the distribution of questions, we grouped pattern- and distance-related categories together, as  particularly in single-room scenarios relevant data is limited.

    \subsection{Taxonomy of Questions}\label{sec:taxonomy}
        
        To achieve a balanced distribution of questions with respect to their types and functions, we investigated the taxonomy of spatial questions primarily proposed for Geographic Information Systems (GIS) by M. Schmidts and N. M. Giner~\cite{understanding2024schmidts}. Based on that, we propose a taxonomy of spatial questions for rooms- and objects-related applications in indoor spaces, whose detailed categories descriptions follow in the corresponding sections. 

        \PAR{Location.} These questions focus on understanding whether objects are present and where they are located, either with respect to the coordinate system or to the associated rooms.
        Some examples of questions in this category are \textit{``Where can you find chairs?''}, \textit{``Are there any sofas in the living room?''}, or \textit{``Which rooms have no plants?''}.
                    
        \PAR{Measurement.} The questions in this category concern providing information related to the size, shape and distribution of both individual objects and rooms. This category includes questions such as \textit{``List all the rooms with the corresponding number of chairs in each of them.''}, \textit{``How many standing lamps are there in all bedrooms?"} or \textit{``Which room has the most sofas?"}.
                
        \PAR{Relation.} The aim of the answers to these questions is to specify the spatial relationships between objects and rooms, summarize what is contained within given areas, and determine what is closest and nearby. Although the original GIS taxonomy proposed to also include the definition of what is visible from a given location, we decided that, in our case, these questions fit better into the prediction category, as we also include a specification of a person's pose. Instead, we suggest replacing it with requests on the descriptions of the room's objects, to transfer the idea of what can be seen in the room. The spatial relationship determination contains the following examples: \textit{``Which objects are within 2 meters from the sofa?''}, \textit{``How are the armchair and the sofa positioned with respect to each other?''}, \textit{``Which rooms are directly accessible from the kitchen?''}.

        \PAR{Navigation.} We associate this group with the objective of finding the shortest navigable or straight-line paths, more specifically, the associated distances in the scene between given rooms or objects \eg \textit{``What is the distance in straight line between the cabinet and the mirror?''}, \textit{``What distance would you have to walk to get from the kitchen to the dining room?''}.

        \PAR{Pattern.} Another aspect of spatial questions covers similarities and patterns identification. Grouping objects or rooms with common features, recognizing spatial and visual trends, and determining consistent or uniform distribution of objects are just some of the topics tackled by the questions in this category. Additionally, we decided to extend this category with the questions on the wall colors, as it requires the feature reasoning based on components (walls) located in different parts of the room or of the apartment. This group includes such questions as \textit{``Which rooms have the same number of chairs?''}, \textit{``Is there a consistent bed distribution across the bedrooms?''}, \textit{``What are the similarities between the two sofas in the living room?''}.

        \PAR{Prediction.} This category of questions requires making predictions based on the room sizes, person's position, and objects' presence and layout. Some examples of these questions are \textit{``How many people could comfortably sit in the dining room?''}, \textit{``What can a person sitting in the chair in the bedroom see in front of them?''}, \textit{``Which room is best adjusted for working on a project alone?''}.
         
    \vspace{-8pt}
    \subsection{Automatic Assessment}
\begin{figure}[tb]
           \centering
           \includegraphics[width=\linewidth]{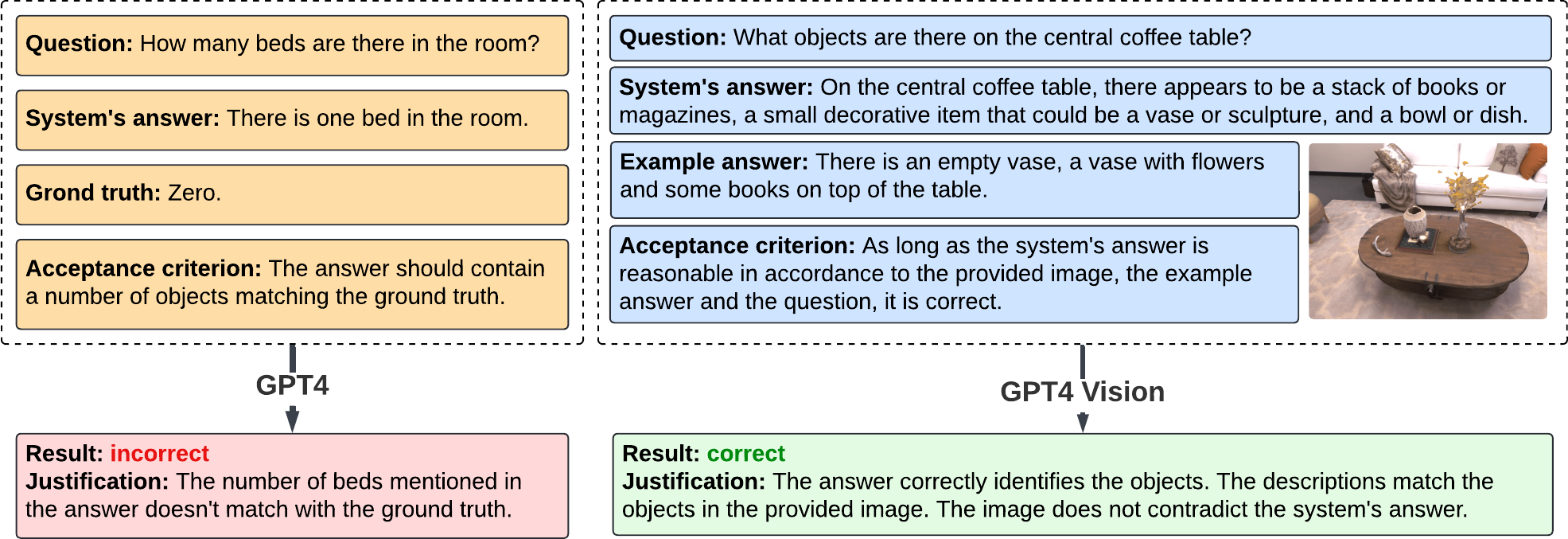}
           \vspace{-15pt}
           \caption{{\bf Automatic assessment procedure.} The left chart presents the scenario of \textit{Ground Truth Check}, the right one depicts \textit{Answer Cross-Check}, used respectively for indisputable data and more creative answers.}
           \label{fig:assessment}
           \vspace{-10pt}
         \end{figure}

        The goal of the automatic assessment is to evaluate the responses from an answering system with respect to the actual state of the corresponding scene in the dataset. The assessment is performed by using LLMs from the GPT4 family as state-of-the-art. We divided the assessment into two cases: \textit{Ground Truth Check} -- when the ground truth is indisputable (\eg number of objects in the room), \textit{Answer Cross-check} -- when the definition of the ground truth would either need to exceed context length or would unnecessarily limit the answering system's creativity (\eg finding similarities between rooms). In both scenarios, the VLM is provided with the question, the system's answer, and the acceptance criterion, which varies based on the question type. In the case of the \textit{Ground Truth Check}, the message to the LLM is extended with information on the actual state of the scene with respect to the given question. \textit{Answer Cross-check}, however, provides an image presenting the corresponding scene(s) in question, accompanied by an example answer. This way, a VLM can decide whether the actual system's answer matches the reality, and not necessarily matching the example, reducing the bias of the assessment system. Then, a language model makes a decision and outputs two components: the acceptance/rejection decision, and its justification, which gives the user of the dataset an insight on what can be addressed in the next iterations of their solution. The examples of the workflows of the assessment system are presented in \cref{fig:assessment}.

\section{Baseline}
\begin{figure}[!tb]
       \centering
       \includegraphics[width=\linewidth]{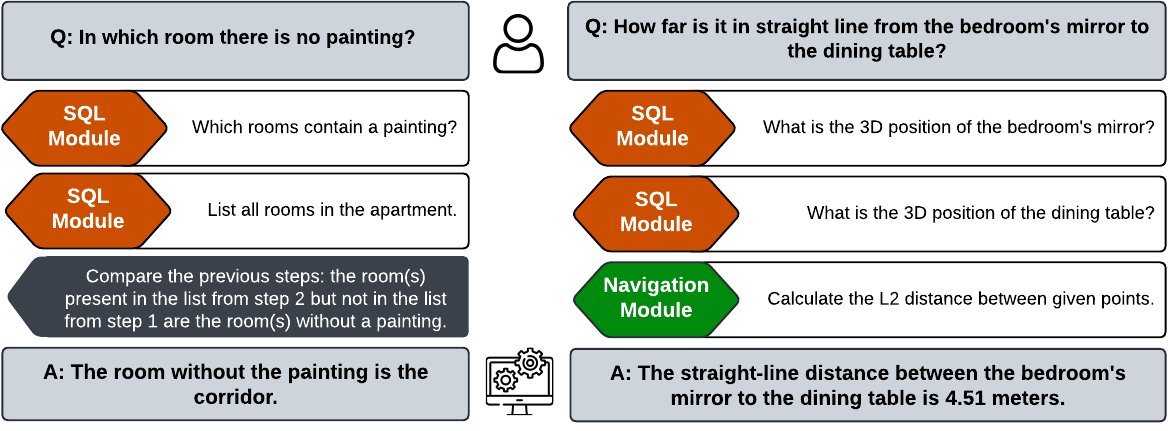}
       \caption{{\bf Example of modules chained together by the planner.} Semantic Kernel's planner divides the user question into subtasks and calls the modules in a sequence to finally formulate a response based on the retrieved information.}
       \label{fig:sk_chain}
       \vspace{-10pt}
     \end{figure}

    To generate answers for the assessment system to evaluate, we propose \baselinename{}, a spatial Q\&A system based on two main components: Semantic Kernel (SK)~\cite{semantic_kernel} and Retrieval Augmented Generation (RAG)~\cite{rag2020} within LlamaIndex framework ~\cite{Liu_LlamaIndex_2022}. Semantic Kernel, being an open-source framework for LLM-based implementation of agents, allowed for integrating four complementary modules -- each with different applications and limitations -- into one system. Once the modules were implemented and described with a corresponding prompt, Semantic Kernel's planner was able to propose a chain of function calls as presented in \cref{fig:sk_chain}, whose result would be an answer to the input question.

    The system requires prior knowledge of the environment in question. Providing it manually by the user would be both tedious and potentially resulting in exceeding the context length. To address this, we implemented Retrieval Augmented Generation pipelines in the LlamaIndex framework for three modules with different sources of data: texts of rooms descriptions, images from the rooms, and an SQL database of detected objects and rooms. The fourth module (navigation) was based on traditional navigation meshes.

\begin{figure}[!tb]
       \centering
       \includegraphics[width=\linewidth]{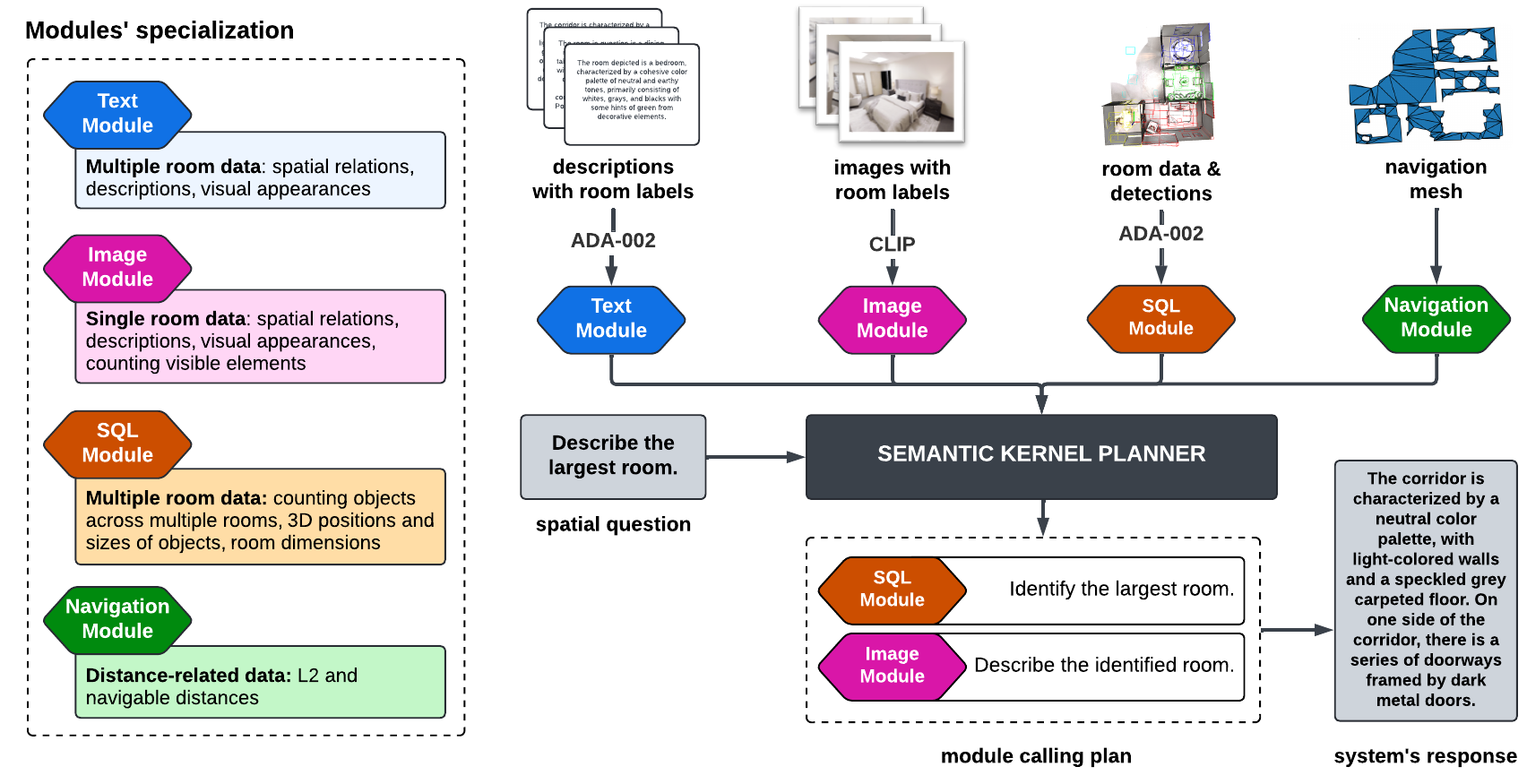}
       \vspace{-22.5pt}
       \caption{\textbf{Overview of \baselinename{}.} Based on the received question, the Semantic Kernel library orchestrates the calls of four different modules having different specializations and types of input.}
       \label{fig:system_overview}
       \vspace{-10pt}
     \end{figure}

    The overview of the system is presented in \cref{fig:system_overview}, with the in-detail explanation of modules present in the following sections.
    
    \vspace{-7pt}
    \subsection{Modules}
        \PAR{Image Module.}
            The Image Module required the collection of RGB images from the dataset, which was done by saving the desired camera frames in Habitat's simulation of each scene. The images were then labelled with corresponding room names. This data was embedded with CLIP~\cite{clip} and loaded into a vector store, with metadata containing the room label.

            Once a question is received by the module, it is embedded with CLIP and passed to a retriever. If an LLM decides that a specific room of the available ones is related to the query, then a metadata filter is added to the retriever to perform a search over the relevant documents only. Then, the image closest to the question in the embedding space is forwarded to a VLM along with the original question for it to provide the final answer.

            As only one image at a time can be provided to the state-of-the-art VLMs (GPT4-Vision~\cite{2023GPT4VisionSC}), this module can only be used in the case of single room data. It can answer questions regarding spatial relationships between objects and their visual appearances, provide room descriptions, or count the number of visible objects.
            
        \PAR{Text Module.}
            The room descriptions required by the Text Module were generated by a VLM. The model was provided with images from two views of each room -- collected similarly to the case of Image Module -- and asked to describe them in great detail, including the visible objects, their visual appearances, spatial relations between them, and the layout of the room. These descriptions were then labelled with corresponding room names. Ready data is embedded with ADA-002~\cite{openai_embedding_model} and loaded into a vector store, with metadata containing the room label.

            As in the case of the Image Module, the question is embedded with the same function as the data in the vector store, and a metadata file is added if specific rooms are mentioned in the query. Then, a query engine performs a search of most relevant data and returns an answer formulated by an LLM. 
            The Text Module complements the Image Module in terms of the application -- it answers similar types of questions, but since more data can be retrieved and passed to the LLM at a time, it is suitable for queries regarding multiple rooms.
        
        \PAR{SQL Module.}
            The SQL Module contains an SQL database with two tables: one for detected objects (with their class names, 3D positions, sizes and associated rooms) and one for information about the different rooms (their names, dimensions and centers).

            The RAG application proved to be useful in the case of SQL query definition. With a limited set of available class and room names in the database, and with the language capabilities of using words from an unconstrained set of synonyms and descriptive phrases, it was necessary to identify both objects and rooms mentioned in the question, and then perform the search of most similar available class and room names in the ADA-002 embedding space. Once these are determined for a specific prompt, the list of top names matches along with the SQL database context is forwarded to an LLM, which creates an SQL query corresponding to the objective of the question. Such a query is then executed and its result is used to formulate an answer with an LLM.

            This module provides quantitative data on the apartment, including object counts across different rooms, room sizes, and object centers.
        
        \PAR{Navigation Module.}
            The Navigation Module in the current state focuses on providing information on both straight-line and shortest navigable distances between rooms and objects. In future work, this could be extended to provide the actual navigation, such as describing the paths between two points.

            The question forwarded to the module must contain explicitly mentioned 3D positions of points between which the distance is to be calculated. An LLM extracts the 3D positions of the questions into a specific format, which makes the text easy to be parsed. In the case of the straight-line distance, the Euclidean norm of the difference of the two positions is computed. In the case of the navigable distance, the navigation mesh is considered. The two points are snapped to the closest points contained by the navigation mesh, and then to the closest vertices. The geodesic distance between those two vertices, if possible to be determined, is output by the module.
    
\vspace{-8pt}
\section{Experimental Results}
    This section is divided into two parts: the user study evaluating the correctness of the automatic answer-assessing system, and the baseline's results on the dataset.
        
    \subsection{Evaluation of the Automatic Assessment}
        To ensure that the automatic assessment correctly accepts or rejects the answers, we conducted a user study. We exposed 60 people to a random sample of 40 questions drawn from a 100-questions scene, asking them to assess the correctness of the system's answers. Additionally, 10 abstracted questions were added to get an insight on people's reasoning in case of ambiguous answers.

\begin{figure}[tb]
           \centering
           \includegraphics[width=\linewidth]{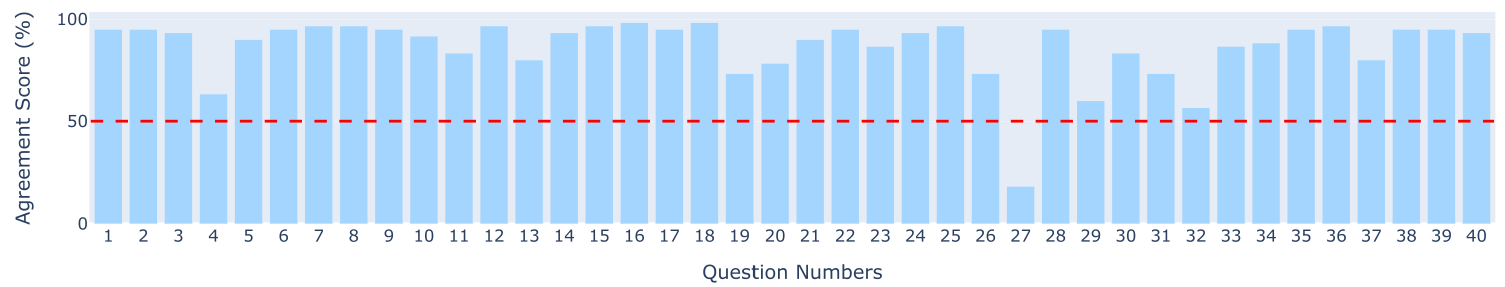}
           \vspace{-20pt}
           \caption{{\bf Results of the user study.} The chart presents the agreement of the participants of the survey with the system. The users agree with the assessment system on 97.5\% of cases, with a weighted agreement score of 86.4\%.}
           \label{fig:user_study}
           \vspace{-10pt}
         \end{figure}

        As presented in \cref{fig:user_study}, the automatic assessment agreed with the majority of participants' responses in 39 out of 40 cases, reaching the absolute agreement rate of 97.5\%. To account for the distribution of participants' decisions with respect to each question, we calculated a weighted agreement score, where the weight for each question was based on the number of users choosing the system-selected option. This metric shows an agreement of 86.4\%, demonstrating a strong alignment with human intuition in the evaluation.
        
        The one question resulting in the disagreement between the system and the majority of people is \textit{``List all the rooms with the corresponding number of chairs in them.''}, while the to-be-evaluated answer contains all rooms apart from the corridor, which does not have any chairs. The automatic assessment does not approve such a response, as the instruction explicitly specifies \textit{all the rooms}. However, participants almost unanimously agree to accept the answer, justifying that the rooms without the objects are not relevant or that even the corridor should not be considered a room.
        
         Certain questions in the user study resulted in the high level of disagreement between the participants themselves. It concerned the cases in which the answering system gave a mostly correct response, but included a minor misalignment with the ground truth. Although the automatic assessment agreed with the majority's decisions, these situations remain ambiguous and should be addressed by a detailed specification of acceptance criteria.

         In addition to the 40 questions on specific answers from the answering system, we included 10 abstracted ones to draw conclusions on how to address ambiguities. For three cases, the participants were almost equally divided between the provided choices, including the decisions on:
         \vspace{-5pt}
         \begin{itemize}
             \item whether an answer about the rooms with \textit{the least} number of objects should consider the rooms without these objects,
             \item if rooms without specified objects have the same number of these objects,
             \item if an answer to a question on the number of objects having the wording \textit{at least} instead of the exact number is correct.
         \end{itemize}
        \vspace{-6pt}
         We believe these divided opinions give a valuable insight on the ambiguities of natural language answers assessment. We release the detailed user study results to support the future research in LLM-based evaluation systems.

         To further analyze the behaviour of the system, we chose Phi3~\cite{phi} as a replacement of GPT4 models. We ran the tests against the questions used in the user study. Phi3 scored 77.5\% of weighted agreement with the participants, failing to align on 7 out of 40 questions. What is interesting to notice, is that the question that for GPT4 caused the disagreement with the users, for Phi3 resulted in the user study alignment. However, the overall performance of Phi3 was much poorer, therefore it was decided to keep GPT4 models as the core of the assessment system until better models are available. 
    \vspace{-7pt}
    \subsection{Baseline Results}
    
\begin{figure}[tb]
           \centering
           \includegraphics[width=0.7\linewidth]{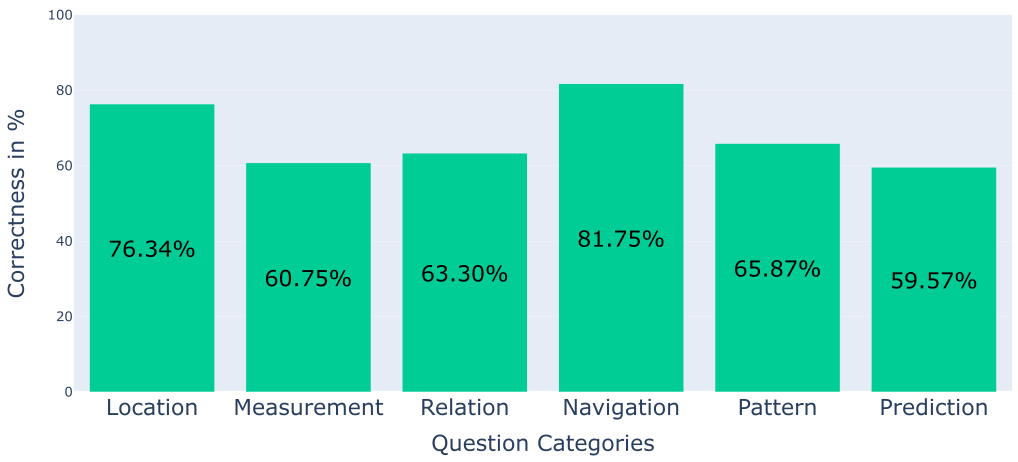}
           \vspace{-10pt}
           \caption{{\bf Results of the baseline on each category of questions.} Predictions pose the biggest challenge for \baselinename{} due to lack of understanding of objects types. Although the highest percentage of correct answers belongs to navigation-related questions, it is the location category with the highest number of correct responses because of the larger number of questions in this group.}
           \label{fig:baseline_res}
           \vspace{-10pt}
         \end{figure}

        The developed system's answers were evaluated by the automatic assessment. The system achieves an accuracy of 66.8\% on the $1000$ questions of the proposed dataset. As presented in \cref{fig:baseline_res}, the system correctly addresses most of the questions in each category. Predictions pose the biggest challenge for the system, accounting for 76 failed cases. When making predictions, the system tends to consider the areas, or sometimes even volumes of the rooms rather than the size and number of the available furniture. It also lacks the understanding of the types of the objects present -- for example, when asked about the number of people that could comfortably sit in the room, it counts the number of seating objects such as chairs, sofas, stools, instead of accounting for the capacity of each. It was additionally noticed that the LLM within the SQL module at times falsely interprets the answers from the SQL database. It is especially manifested in the case of counting the number objects which are not present in the scene -- occasionally the returned zero is interpreted as an object existence. The reason for that particular phenomenon can be investigated in the future work.

    \vspace{-7pt}
\section{Limitations \& Future Work}
    We believe that our dataset is a valuable asset in evaluating spatial Q\&A systems. However, various challenges remain open. The questions in the distance-related category in the dataset could be extended with navigation aspects, such as describing paths to be followed by a person to get from one place to another. It would also be valuable to get an insight on whether answer formulation affects the assessment system. For example, in the prediction regarding best rooms for a specific activity, instead of providing only the names of the rooms that seem a reasonable choice to us, humans, a description of relevant properties of each room could be provided for the assessing VLM to make a decision and, perhaps, give more slack to the answers. Moreover, as Replica offers a relatively small number of scenes, it would be beneficial to extend the Q\&A to other datasets, ideally with multi-floor, multi-room scenes.

    As natural language tends to be ambiguous, a more detailed user study should be conducted, to determine how human intuition perceives vague or partially-correct answers. Although we included a couple of abstracted questions in our survey, more research in the area would prove useful for future Q\&A datasets.
    
    In terms of auxiliary data we provide, the curated Replica object detections could be improved. Object-to-room assignments, for instance, assume rectangular, axis-aligned properties of each room, which proved to be an acceptable approximation. For the sake of higher accuracy and reliability, a better assignment procedure should be employed.

    There are a couple of known limitations of the baseline that should be addressed in the future work. The most evident one is the lack of a bridge between the objects' semantics (their colors, shapes, types), and their quantitative data (3D positions, sizes, number of objects). It would be interesting to experiment with integrating object descriptions into the SQL database, potentially with RAG implementation on relevant objects' row-retrieval. It would also be valuable to investigate if and how rephrasing the questions would affect the performance of the answering system.
\vspace{-8pt}
\section{Conclusions}
    In summary, this paper presents a dataset of $1000$ question-answer pairs for spatial Q\&A, based on different indoor scenes with a variety of modalities available. The dataset is balanced with respect to the proposed spatial question taxonomy, initially employed in geography-related research. The benchmark is complemented with an answer-assessing system, which leverages a VLM's text and image understanding. The assessment's correctness is justified by the results of a user study. The proposed baseline system applies Retrieval-Augmented Generation with multiple chained VLMs to address spatial Q\&A, and achieves 67\% accuracy on the proposed dataset. Although the baseline efficiently handles a variety of question types, there is room for improvement in its 3D reasoning and scene understanding.

\section*{Acknowledgements}
    We thank all the participants of the user study for their time and effort. Special appreciation goes to our colleagues from Microsoft, the ETH CVG lab, and those who joined us through the authors' social media channels.

\bibliographystyle{splncs04}
\bibliography{main}
\end{document}